# A Multivocal Literature Review on Privacy and Fairness in Federated Learning

Beatrice Balbierer[1], Lukas Heinlein[1], Domenique Zipperling[1,2], Niklas Kühl [1,2]

[1] University of Bayreuth, [2] Fraunhofer FIT
{beatrice.balbierer, lukas.heinlein, domenique.zipperling, niklas.kuehl}@uni-bayreuth.de

**Abstract.** Federated Learning presents a way to revolutionize AI applications by eliminating the necessity for data sharing. Yet, research has shown that information can still be extracted during training, making additional privacy-preserving measures such as differential privacy imperative. To implement real-world federated learning applications, fairness, ranging from a fair distribution of performance to non-discriminative behavior, must be considered. Particularly in high-risk applications (e.g. healthcare), avoiding the repetition of past discriminatory errors is paramount. As recent research has demonstrated an inherent tension between privacy and fairness, we conduct a multivocal literature review to examine the current methods to integrate privacy and fairness in federated learning. Our analyses illustrate that the relationship between privacy and fairness has been neglected, posing a critical risk for real-world applications. We highlight the need to explore the relationship between privacy, fairness, and performance, advocating for the creation of integrated federated learning frameworks.

**Keywords:** *Federated Learning, Machine Learning, Fairness, Privacy*

## 1  Introduction

In today's digital age, the increasing availability of data offers unprecedented opportunities for innovation, particularly for machine learning (ML) (Courville et al. 2018). However, utilization remains a challenge as data is usually not stored centrally, but distributed across data owners, i.e., clients, like edge devices or organizations (Zheng et al. 2023; Wang et al. 2022; Hosseini et al. 2023). Yet, storing data centrally is infeasible due to privacy and intellectual property (IP) exposure risks (Wieringa et al. 2021). Federated Learning (FL) can inherently mitigate these privacy risks by avoiding the centralization of sensitive data (Mothukuri et al. 2021; Wen et al. 2023). To harness all data, clients utilize their data to train an ML model locally and only transfer model updates to a central server, aggregating these updates to a final model (McMahan et al., 2016). By not sharing raw data, FL has opened up new possibilities for ML applications for edge devices (Saylam und İncel 2023) or more sensitive domains such as healthcare

(Joshi et al. 2022). Although the centralization of data is avoided, the risk of information leakage remains as shared updates can be analyzed to extract information about the training data, thereby posing a privacy and/or IP exposure risk (Wei et al. 2020; Zhang et al. 2020). Additionally, aggregating multiple local model updates into a single final model that is then utilized by all clients can lead to unequal performance across clients or disparate outcomes for underrepresented social groups (Su et al. 2024) This is a common issue when dealing with machine learning models based on historical data, often referred to as algorithmic fairness (Ledford 2019; Shi et al. 2021). Applications of FL in healthcare illustrate the importance of privacy and fairness quite vividly as patient data, due to its sensitive nature, must be kept private. Additionally, the application's performance, e.g., detection of skin cancer (Kumar et al. 2024), should neither vary across hospitals nor ethnic groups as it is not the case among physicians (Fahmy et al. 2023). Unfortunately, achieving both objectives simultaneously is not trivial and measures to ensure privacy may negatively affect efforts to ensure fairness (Gu et al. 2022). While privacy has been a focus within FL research for quite some time, less focus has been on fairness demonstrated by the lack of fairness aspects in recent surveys regarding current challenges in FL (Guo et al. 2024; Guendouzi et al. 2023; Wen et al. 2023).

With our research, we want to explore current methods that enhance privacy or foster fairness in FL while especially focusing on efforts to fuse the two as they contribute to the development of FL systems that are both technologically sophisticated and ethically sound. To evaluate the current methods, we will explore the following research question (RQ):

**What are current methods that ensure privacy and fairness in federated learning independently and together?**

Using a multivocal literature review (MLR), based on Kitchenham und Charters (2007) guidelines for conducting a structured literature review, extended by the inclusion of gray literature as suggested by Garousi et al. (2019), we assess current methods for privacy and fairness in FL. Choosing an MLR instead of a strategic literature approach allows us to capture literature that is not formally published and thereby might capture emerging research topics, which is especially relevant when analyzing the current body of knowledge in fast-evolving fields like ML and FL.

Our findings highlight that research is centered around avoiding data leakage, and ensuring data integrity while fostering fairness regarding performance across clients (client fairness) and social groups (group fairness). The identified literature demonstrates that the relationship between privacy and fairness has been neglected. Furthermore, we find that over half of the identified literature does not focus on certain domains or explicit applications but rather on developing new methods. Therefore, the results of our work encourage and call for research at the intersection of privacy and fairness with a focus on real-world applications.

## 2 Conceptual Background

The conceptual background provides a basic understanding of FL and its relation to privacy and fairness. Therefore, this chapter explores the mechanisms of FL, while defining the terms of privacy and fairness more closely.

FL is an innovative paradigm for enabling decentralized training of ML models (McMahan et al. 2016). Training takes place on distributed devices or clients, eliminating the need for centralized storage of raw data. Thereby, FL protects privacy as the data remains on individual clients contributing to the improvement of the model by sharing model updates (Mothukuri et al. 2021). As a result, FL facilitates the cross-company use of AI and improves data access without disclosing personal data, which is crucial for exploiting the full potential of AI in the industrial sector (Feuerriegel et al. 2024). This not only improves data protection but also enables flexible adaptation to the different operating conditions of different companies. The typical process includes initializing a global model, distribution to devices or clients, independent training with local data, aggregation of updates, and iterative refinement of the model. FL is used in areas such as healthcare, finance, and edge computing to address privacy concerns while enabling collaborative model building. This approach represents a technical advance and marks a shift toward decentralized, privacy-centric ML methods (Rieke et al. 2020). Although FL avoids sharing raw data, privacy concerns are not muted. Privacy within information systems is usually connected to information and definitions "typically include some form of control over the potential secondary uses of one's personal information" (Bélanger und Crossler 2011). Various attacks such as membership inference attacks can affect the control over potential use by extracting information about the training data based on the shared model updates, thereby compromising privacy (Zhang et al. 2020). This is often referred to as data leakage (Wang et al. 2019; Ren et al. 2022a).

While privacy has been a focus within FL research for quite some time, less focus has been on fairness which is demonstrated by the lack of fairness aspect in a recent survey regarding the current state and challenges in FL (Guo et al. 2024; Guendouzi et al. 2023; Wen et al. 2023). Yet, fairness is crucial when designing inclusive and efficient FL applications (Shi et al. 2021). To address this challenge, Su et al. (2023) emphasize the importance of two key dimensions of fairness: group and client fairness. Client fairness focuses on equal distribution of performance and fair compensation for clients' contribution of data or computational resources. Group fairness refers to debates about algorithmic fairness and the mitigation of biases towards underrepresented social groups.

## 3 Method

Our study aims to provide a comprehensive overview of the current methods to ensure fairness and privacy in FL applications and to identify future research directions. Therefore, we conducted a multivocal literature review (MLR) following Kitchenham und Charters (2007) guidelines for academic literature (AL), supplemented by the inclusion of gray literature (GL) to ensure a more holistic perspective by including not formally published literature. As recommended by (Garousi et al. 2019) including gray literature is especially useful to identify emerging research topics. As the field of FL and ML in general evolves rapidly, including gray literature is important to capture all relevant topics. Choosing an MLR facilitates the identification of a significant number of relevant articles and the development of a concise dataset, supporting the identification of further research needs.

For AL, we screened eight databases covering various interdisciplinary fields using the search string, "Federated Learning" AND (Priva* OR Fair*), yielding 2,746 hits. We defined inclusion and exclusion criteria (Table 1) and performed a two-step title screening, assessing the relevance and contextual contribution of the articles individually and then collectively. In addition, a stop criterion was defined to include only the first eight pages of a database if it does not offer an export function. To enhance the relevance of the publications in our database, we analyzed the citation rates of the articles using the importance of citations as described by Kladroba et al. (2021). Specifically, for articles published until 2022, a minimum of three citations is required, while works from 2023 must have at least one citation.

**Table 1.** Inclusion and exclusion criteria

| Inclusion Criteria | Exclusion Criteria |
| --- | --- |
| Availability of the full text. | Brief mention of the research content without significant contribution to the existing knowledge. |
| Publication before 2024. | The mere outline of the research question without in-depth analysis or discussion. |
| Paper's relevance to the research question. | Privacy-preserving methods and fairness examinations unrelated to federated learning approaches |
|  | Federated learning is employed as the methodology for ensuring the preservation of privacy. |
|  | Papers published at MDPI due to concerns regarding the peer-review process |

The papers underwent relevance evaluation by the author team using a 1-to-5 Likert scale, where 5 indicates high relevance, 3 denotes neutrality, and 1 reflects irrelevance. A rating of 5 was assigned to papers that made a direct and significant contribution to the field of research. A rating of 4 implied that the paper contained relevant information but was less central. A score of 3 indicated general relevance but without direct contribution to the core of the research. Only papers scoring 4 or above were included in the dataset. Any disagreements were resolved through discussion, leading to 62 relevant AL items. Regarding the GL, the identical search terms and stop criterion were used in Google, Google Scholar, and arXiv to ensure consistency in our methodology. We choose the sources in alignment with the methodology and other papers performing an MLR (Paez 2017). The preliminary search yielded 488 GL items of GL. However, these were subject to a rigorous evaluation process, employing the same inclusion and exclusion criteria used for AL. Following this evaluation, only eleven GL items were deemed relevant and of sufficient quality to be included in the final dataset. Furthermore, snowballing was performed and evaluated using the previously defined criteria, with the object of exhausting all literature sources. This yielded 38 additional relevant papers, bringing the total number of items in our final dataset to 133 (see Figure 1).

Our research aims to examine current methods in the area of privacy and fairness in the context of FL. To this end, a two-stage analysis process was undertaken to provide a comprehensive overview of these issues. The first phase focused on privacy in the context of FL, followed by a detailed consideration of fairness approaches in this area. Each section of the analysis presents the technologies identified and their interrelationships with other technologies. The goal is to provide a comprehensive overview of the most common approaches and techniques, thereby deepening the understanding of these key areas.

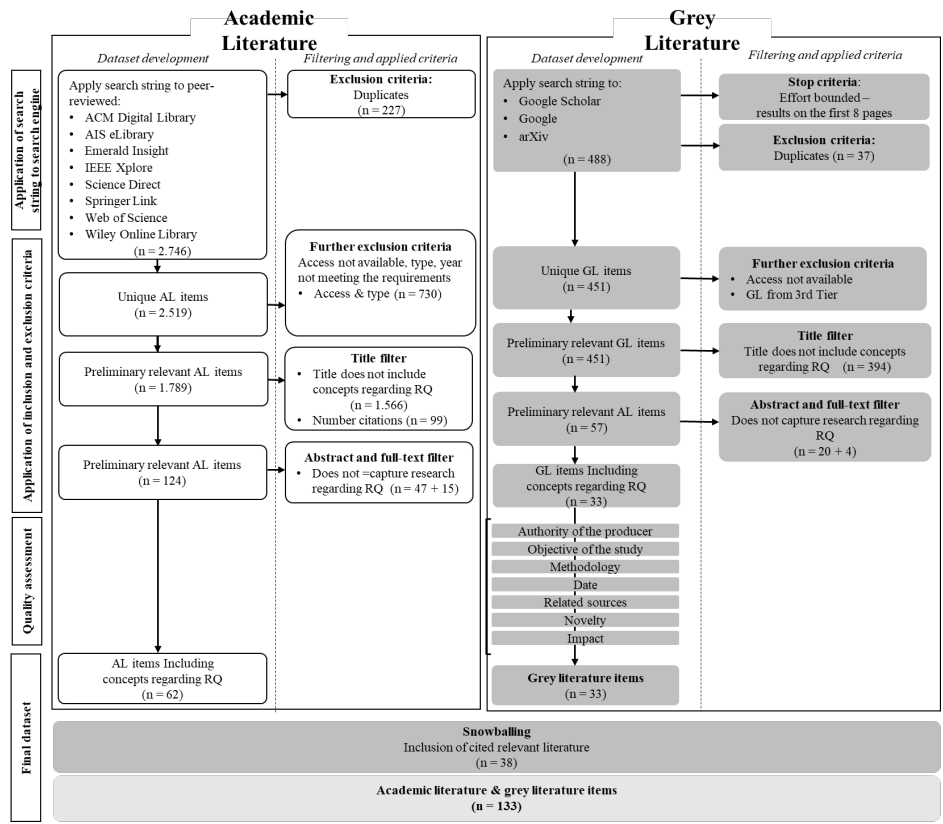

**Figure 1.** Search and selection process

## 4 Results

To capture the current state of research on fairness and privacy in FL, we examine the characteristics of the identified literature. We analyze the distribution concerning literature classification and investigate specific sub-dimensions of fairness and privacy. Our goal is to pinpoint where recent research has concentrated its efforts and thereby identify which dimensions, sub-dimensions, and cross-dimensional inquiries have been neglected.

### 4.1 Characteristics of the included publications.

A review of eleven databases was conducted to identify relevant literature on privacy and fairness. A total of 45 significant papers were found in Web of Science, 15 in IEEE Xplore, 8 in ScienceDirect, 7 in ACM Digital Library, 5 in SpringerLink, and 4 in Wiley Online Library. ArXiv had the highest number of publications in the field of GL with 24 entries, followed by Google Scholar (23) and Google (2). Analyzing the thematic distribution of articles, there is a strong overrepresentation in the privacy dimension, with 98 articles addressing privacy concerns and only 24 articles discussing fairness issues. Privacy research is predominantly associated with AL (70%), while fairness research is almost equally split between AL and GL. Only 11 papers examine both privacy and fairness simultaneously (four from AL and seven from GL). We identified two sub-dimensions for each dimension: data leakage and data integrity for privacy, and client and group fairness for fairness. Data leakage has attracted the most attention, with 72 papers addressing it alone and another 35 in combination with data integrity. Data integrity specifically is addressed in two papers. Fairness research focuses on emphasizing client fairness (23 papers) while group fairness is addressed in eight papers. Four papers address both dimensions of fairness. Cross-dimensional research is scarce, with eleven articles addressing at least one privacy and one fairness dimension simultaneously. No literature was identified that covers both dimensions of privacy and both dimensions of fairness simultaneously.

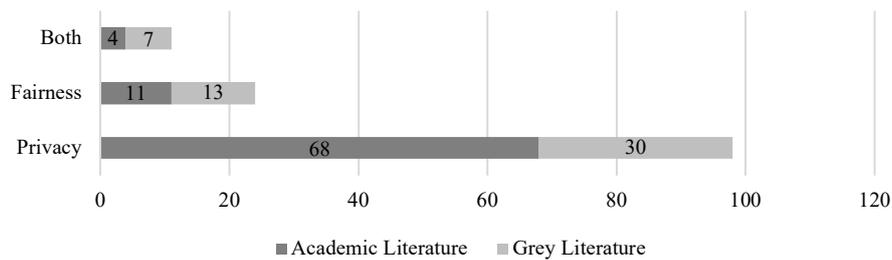

**Figure 2.** Thematic distribution of publications

## 4.2 Overview of methods to address privacy and fairness in federated learning

After describing the characteristics of the included literature, we will overview different methods to ensure privacy or fairness, including brief descriptions. We will also illustrate how these methods are combined to achieve various objectives within or across dimensions. Figure 3 shows the number of papers focusing on one or both dimensions and the specific sub-dimensions they address. Additionally, the figure provides insight into the applications: the first number shows the papers that are domain-unspecific, and the second shows those focusing on domain-specific applications. A detailed description of the domains is provided in chapter 4.3.

**Methods used to ensure privacy.** Preventing data leaks and maintaining system integrity are crucial challenges in the field of FL. A variety of different methods are used to protect user data from disclosure by anonymization or encrypting. The most common techniques include Differential Privacy (DP) with 56 papers, Homomorphic Encryption (HE) with 34 papers, and Secure Multi-Party Computation (SMPC) with 17 papers. DP adds noise to the actual dataset thereby enabling the extraction of aggregated or static information from a data set while protecting the privacy of individuals (Li et al. 2023) addressing privacy concerns in industries like healthcare. Exemplarily, DP is employed in Label DP (Tang et al. 2023) or combination with stochastic gradient descent for secure aggregation. Label DP is used in classification tasks obscuring the direct link between features and their labels while DP for secure aggregation preserves the privacy of model updates when shared with a server for model aggregation (Ghazi et al. 2021). HE prevents data leakage by enabling computations to be performed directly on encrypted data without prior decryption or reduction in the liability of the result (Lloret-Talavera et al. 2021). 17 papers are identified combining DP with HE providing two levels of privacy: a user and a model-based level. Another approach is to combine HE with secret sharing (Shi et al. 2023). Secret sharing divides information into several parts, called shares. The shares themselves do not reveal any insights and the information can only be reconstructed by combining a certain number of shares. Thereby, greater flexibility and scalability are offered as new shares can be added and removed without changing the entire system while increasing data security. SMPC enables multiple users to pool their data and perform computations without disclosing their private information to each other while cryptographic techniques ensure the correctness of the final result (Zhu 2020). All identified papers consistently use SMPC in conjunction with DP enhancing both privacy preservation and data usability. SMPC enables secure computation, while DP adds noise to the data to ensure user anonymity (Abaoud et al. 2023). Another layer of security can be provided by applying HE to a combination of SMPC and DP, as introduced by seven identified papers. The combination of these three methods results in a robust data protection system that ensures the secure exchange and analysis of information while maintaining privacy (Kadhe et al. 2023). Yang et al. (2023) and Maurya und Prakash (2023) introduce specialized approaches that use a particular form of SMPC called secure aggregation to complement the application of DP and HE.

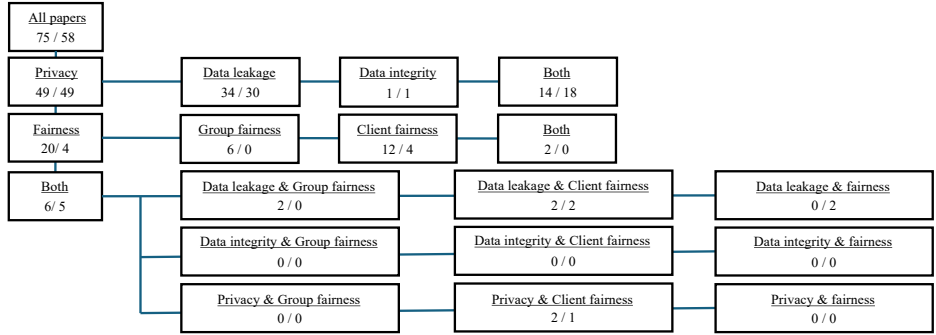

**Figure 3.** Overview of the identified literature grouped by (sub-) dimensions and whether there is a focus on an application domain (e.g., healthcare). For each (sub-) dimension, the first number represents the number of papers without focusing on an application domain (domain-unspecific), while the second number represents the number of papers with an application domain (domain-specific).

Secure aggregation is specifically designed for the aggregation of FL systems, while SMPC has a much broader application area (Bonawitz et al. 2016). Smahi et al. (2023) propose an approach that emphasizes privacy protection to ensure data integrity. The approach primarily focuses on using FL to protect privacy while utilizing BC and zero-knowledge proofs (ZKPs) as sub-elements. Thereby, model updates are stored securely in an immutable and transparent manner while keeping the actual data confidential. BC-based systems therefore generally represent a technological advance, as these decentralized ledger technologies contribute significantly to the transparency and reliability of digital transactions and foster trust in decentralized processes (Hayati et al. 2022). Our analysis of the literature shows that BC, complemented by other techniques, is oftentimes utilized to address data leakage as well as data integrity (Kalapaaking et al. 2023). The Federated Visual Learning System, known as FedVisionBC, mitigates issues such as single point of failure, model poisoning, and membership inference attacks (Zhang et al. 2023b). Consensus protocols such as Delegated Proof of Stake and Proof of Authority strengthen the security mechanisms and trustworthiness of decentralized systems (Fang et al. 2022, Ullah et al. 2023). Recent advancements in secure FL for IoT and edge computing include the integration of Learning with Errors encryption and Consistent Hashing Algorithm for secure data computations (Ren et al. 2022b) and a model parameter aggregation protocol that ensures data privacy on user devices introduced by Eltaras et al. (2023) Verifiable Privacy-Preserving Federated Learning for edge computing systems uses HE and combines Distributed Selective Stochastic Gradient Descent with the Paillier cryptosystem to achieve efficient distributed encryption (Zhang et al. 2023a). Dynamic Successive Verification Mechanisms use HE thereby enhancing security and privacy through dynamic verification and encryption (Gao et al. 2023a). This protocol, based on Boneh-Lynn-Shacham signatures and multi-party security, verifies the integrity of client-uploaded parameters and the correctness of server-aggregated results. Additionally, approximate homomorphic Cheon-Kim-Kim-Song encryption protects client data privacy (Ma et al. 2022).

**Methods used to achieve fairness.** In the area of FL, there are a variety of research efforts to improve both fairness and accuracy. A wide range of innovative approaches and technologies are being explored, including developing personalized models, innovative aggregation methods, and asynchronous mechanisms. The papers mainly present use cases related to inter-hospital collaboration and learning from histopathology images (Hosseini et al. 2023, Li et al. 2019, Zhao und Joshi 2022). Client fairness in FL aims to evenly distribute benefits among clients, focusing on equitable performance distribution, resource-driven participation, collaboration incentives, trust assessment, and reducing bias through algorithmic fairness. A key approach involves crafting personalized models tailored to the unique needs and data profiles of each client, essential for fostering more equitable and impactful outcomes. Innovative aggregation methods, such as the use of double momentum gradients, can enable more efficient and equitable integration of contributions from various clients (Huang et al. 2022). Asynchronous mechanisms address the variability in client availability and computing power by allowing clients to submit updates on their schedules without the need for synchronized communication. This approach leads to more efficient resource utilization and ensures fairer participation among clients (Gu und Zhang 2023). Another approach uses incentivization mechanisms, such as reverse auctions and trust assessments, to promote efficiency and ensure a fair distribution of resources (Pan et al. 2023). Proportional Fair FL adjusts weights dynamically addressing the challenge of unequal starting conditions of clients (Hosseini et al. 2023). Multi-gradient descent and adaptive accuracy control mechanisms enhance the efficiency and fairness of the methods used by allowing models to learn in different directions and dynamically adjust learning rates and other parameters to achieve optimal results under varying conditions (Fu et al. 2020).

Group fairness is addressed by five articles from GL and one from AL introducing six different approaches: FedGFT (Wang et al. 2023), cCFLVvis (Huang et al. 2023), Fair-Fate (Salazar et al.), multiparty calculation (Pentyala et al. 2022), FedMinMax (Papadaki et al. 2022), and FedFB (Zeng et al. 2021). These approaches aim to reduce systematic discrimination towards certain population groups. Fair-Fate adjusts the weighting of clients based on their contribution to the overall model, providing a fairer distribution of performance results across subgroups defined by sensitive characteristics (Salazar et al.). In contrast, FedGFT is designed to achieve global fairness by regulating an objective function that balances individual performance and collective benefits (Wang et al. 2023). This approach promotes demographic fairness and mitigates the Matthew effect (Gao et al. 2023b). Techniques such as dynamic reweighting and adaptive accuracy checks can promote a more even distribution of performance, creating a more inclusive and equitable learning environment (Zhao und Joshi 2022). Two identified papers cover both, client and group fairness. Fair Hypernetworks (Carey et al. 2022) adjust the parameters of a target network based on various fairness criteria. They are trained to minimize discrimination and promote fair decisions between different groups, aiming to balance model performance and fairness (Qu et al. 2022). Dynamic Q Fairness Federated Learning Algorithm with Reinforcement Learning (Chen et al. 2023) combines FL with reinforcement learning methods to improve both fairness and efficiency. The algorithm uses a dynamic Q-function to make optimal decisions for model updates and resource allocations, while continuously adapting to changing conditions (Cao et al. 2024). This maximizes the overall performance of the federated learning system while ensuring that all participating devices are treated fairly (Woo et al. 2023).

**Methods to address privacy and fairness.** Eleven papers were identified that consider both privacy and fairness in their analysis. The papers present a variety of methods to improve fairness, privacy, and efficiency in FL, covering the use of cryptographic methods, as well as the development of specific algorithms and frameworks. You et al. (2023) and Ratnayake et al. (2023) cover data leakage and client fairness. While Ratnayake et al. (2023) present a review that examines and extends the existing taxonomy of fairness approaches in FL and analyzes the implementations of different methodological groups, the other three papers discuss distinct methodological approaches. You et al. (2023) introduce the FedACC framework, a server-led approach for controlling the global model accuracy. It ensures the validity of client gradients, and measures the cumulative contributions of new clients, while only granting access to a model if the accuracy matches their contributions. Utilizing differential privacy adds a layer of privacy. Annapareddy et al. (2023) address both fairness dimensions (group and client fairness) while additionally considering data leakage by introducing the FedFa using a double momentum mechanism to process non-IID data. The server aggregates client information by considering both historical gradient information and quantity information about client accuracy and participation frequency. An appropriate weight is determined for each client based on this information. Furthermore, Rückel et al. (2022) cover both dimensions of privacy (data leakage and integrity) while addressing client fairness by introducing a BC-based architecture for FL systems that combines ZKPs, local DP, and BC to create a fair, and manipulation-proof FL environment. This system allows clients to verify the integrity of each other's model training, promoting fair compensation. The novel use of ZKPs ensures both accurate model inferences and fairness throughout the training process.

## 4.3 Overview of application domains and technology-focused research

While identifying methods to ensure group and client fairness, and data integrity, and mitigate data leakage is important, it's crucial to analyze their application to specific use cases. About 60% of the literature focuses on theoretical and initial empirical evidence without specifying an explicit domain. The applied research is mainly centered around healthcare, IoT, and edge devices. The aggregated data shows that Healthcare (8 privacy, 1 fairness, 1 both), IoT & Edge Computing (15 privacy, 0 fairness, 2 both), and Big Data & Cloud Computing (6 privacy, 2 fairness) are frequently discussed in terms of privacy concerns. Privacy research is almost equally divided between domain-specific and domain-unspecific studies, while fairness research is mostly domain-unspecific. Categories such as Cyber-Physical & Autonomous Systems (3 privacy, 1 both), Finance (4 privacy, 1 both), Industry & Communication (7 privacy), and Energy Sector (1 fairness) show broader applications. Mobility & Traffic (2 privacy) and Multimedia & Visual Detection (4 privacy) also focus more on privacy. This indicates a need for more application-specific fairness studies to match the developed state of privacy research. Future research should bridge this gap by developing and validating theoretical frameworks in practical applications across various sectors.

## 5  Discussion

Using a multivocal literature review, we identified 133 papers surrounding privacy and/or fairness in federated learning. The results show a clear focus of research on privacy (98) and demonstrate the lack of focus on fusing privacy and fairness concerns as only 11 papers were identified that look at (minimum) one privacy and one fairness dimension simultaneously. When it comes to privacy, there were three core methods identified: differential privacy, homomorphic encryption, and secure multi-party computation. Interestingly, the identified research regarding privacy also focuses on security with an emphasis on the integrity of data (35 papers). Here, one or multiple mentioned technologies are combined with blockchain technology. Research on fairness (24 papers) is mostly centered around client fairness (16 papers) while six papers focus on group fairness and only four papers consider both dimensions. Additionally, almost half of the research on privacy focuses more on the implementation of methods in real-world applications. In contrast, only 17% of research regarding fairness focuses on specific domains. Especially research on group fairness has no connection to specific domains, indicating that research on (group) fairness in FL is mostly abstract and still in its infancy as literature focuses on the development of methods rather than their implementation in the real world.

The lack of fairness-centered research, especially group fairness, is concerning in light of newly passed legalization like the AI Act (European Comission 2024) calling for the detection and mitigation of biases during the development and deployment of machine learning-based applications. Furthermore, as current legislation on privacy laws like the General Data Protection Regulation (GDPR) (*General Data Protection Regulation.* European Council, 2024) put high demands on the privacy of data combined with known trade-offs between dimensions, e.g., anonymization for privacy vs. ensuring group fairness, research must focus on the interplay between different methods (Kusner et al. 2017, Dwork et al. 2011). Therefore, we call for the development of a holistic FL framework considering the dependencies across dimensions.

To achieve this goal, we propose the following research directions. Firstly, we call for interdisciplinary research incorporating technical, ethical, legal, and social considerations into the design and deployment of FL applications. Initial research must extend the knowledge of trade-offs between the dimensions on an empirical level. Inspired by (Gu et al, 2022) we propose to analyze the interplay between two sub-dimensions, and how they impact each other. Furthermore, we are interested in how they could influence model accuracy in a controlled environment. To keep complexity managed, future research should focus first on understanding relationships between two sub-dimensions more deeply. In the next step, research should push toward understanding the dependencies between data leakage, data integrity, client fairness, and group fairness. Additionally, research should focus on implementations in real-world scenarios thereby leaving the laboratory-like evaluation. Here, a design science research approach could help identify suitable design requirements, principles, and features which is especially true for fairness concerns.

Furthermore, we want to highlight a neglected aspect of the privacy-preservation debate in FL: the right to be forgotten, a key component of data privacy laws. Despite some integration efforts into FL, our findings indicate it has been overlooked in current discussions. Ignoring such a crucial right could hinder the real-world application of FL. Additionally, we advocate for a life-cycle-centric approach to ensure group fairness. Fairness must be considered holistically, not just during model training (Deck et al. 2024). Furthermore, focusing on additional fairness dimensions will help to gain a holistic perspective. A prominent candidate could be informational fairness (Schoeffer et al. 2022). Lastly, we highlight a managerial challenge in ensuring group fairness in FL. Group fairness relies on a normative definition that influences the choice of fairness metrics and is already challenging in centralized ML applications (Saleiro et al. 2018; Franklin et al. 2022). How independent clients, such as organizations or companies, can agree on a single definition of fairness and its metric remains a major challenge for FL to become practical.

# 6   Conclusion

This paper provides a comprehensive overview of the methods currently presented and discussed for ensuring privacy and fairness in federated learning systems. For this purpose, an extensive multivocal literature review was conducted to highlight recent research efforts, summarize methods to ensure privacy and fairness and identify research directions. Subsequently, we identified sub-dimensions on which privacy and fairness research has been focused: data leakage, data integrity, client fairness, and group fairness. This was followed by a detailed discussion of methods in each (sub-)dimension and whether there have been efforts to combine them. Furthermore, application areas are discussed illustrating that the focus of research has been on applying federated learning in healthcare. Here, ensuring privacy and fairness are paramount for real-world applications. We discussed the challenges of integrating privacy and fairness in federated learning to identify further research avenues for future research. Among them is more research in the trade-off between multiple dimensions, implementing them in more real-world scenarios while utilizing the design science research paradigm. Furthermore, missing components such as unlearning or informational fairness were proposed to illustrate further research directions. Yet, our research is still limited as it only provided a superficial analysis of methods. In future work, the collected data can be used to create a more detailed picture of methods, their precise intentions as well as their technological readiness. Nevertheless, we hope our findings will serve as a valuable roadmap towards fairer and more private applications for federated learning.